\documentclass[11pt]{article}

\usepackage[preprint]{acl}

\usepackage{times}
\usepackage{latexsym}

\usepackage[T1]{fontenc}
\usepackage[utf8]{inputenc}

\usepackage{microtype}

\usepackage{inconsolata}

\usepackage{graphicx}

\usepackage{booktabs}
\usepackage{tabularx}
\usepackage{multirow}
\usepackage{pgf}
\usepackage{pgfplots}
\usepackage{subcaption}
\usepackage{amsmath}
\usepackage{enumitem}

\title{Complete Evidence Extraction with Model Ensembles: \\A Case Study on Medical Coding}

\author{
 \textbf{Katharina Beckh\textsuperscript{1,2}},
 \textbf{Sven Heuser\textsuperscript{1}},
 \textbf{Stefan Rüping\textsuperscript{1}}
\\
 \textsuperscript{1}Fraunhofer IAIS\\
 \textsuperscript{2}Lamarr Institute
\\
 \small{
   \textbf{Correspondence:} \href{mailto:katharina.beckh@iais.fraunhofer.de}{katharina.beckh@iais.fraunhofer.de}
 }
}

\begin{document}
\maketitle
\begin{abstract}
High-stakes decisions informed by decision support systems require explicit evidence.  
While prior work focuses on short sufficient evidence, regulatory compliance and medical billing call for \emph{complete} evidence: all relevant input tokens that support a decision. 
We formulate complete evidence extraction as a task and study it in a medical coding setting. 
Motivated by the Rashomon effect, we aggregate token-level evidence from multiple language models to increase evidence completeness. 
We perform a case study using existing equally-performing models, feature attributions, and a dataset with human-annotated evidence. 
Our results show that Rashomon ensembles significantly increase evidence recall while incurring only a small token overhead over individual models. Ensembles of only three models already outperform the best single model and recover information that individual models miss. 
\end{abstract}

\section{Introduction}
In many high-stakes settings, stakeholders must be able to justify decisions. As decision support systems are more widely used for classification, these evidentiary requirements extend to the systems themselves.  
Evidence extraction methods address this need by identifying input features that support a model prediction.\footnote{Evidence does not necessarily provide insight into \textit{how} a prediction was reached \cite{tan-2022-diversity}, hence we use the more specific term evidence instead of explanation or rationale.} 
Most prior work focuses on minimal \textit{sufficient} justifications \cite{lei-etal-2016-rationalizing, thorne-etal-2018-fever,luss2024when}, where evidence is short: often a single word or phrase at one position in the text.

However, in critical settings, such as regulatory compliance~\cite{beckh2024limitations,schmitz2024towards} and clinical medicine~\cite{noll2023new, li2024a}, stakeholders need \textit{complete} evidence, which identifies all possible tokens at different positions in the document.
For example, in psychiatric care, the number and severity of indicators, such as self-harm and aggression, determine the need for hospitalization and impact billing \cite{noll2023new}. If a decision-support system fails to retrieve relevant indicators and the threshold for an extended stay is not met, a patient might be discharged from the clinic too early -- a risk for the patient and others.

\begin{figure}
    \centering
    \includegraphics[width=\linewidth]{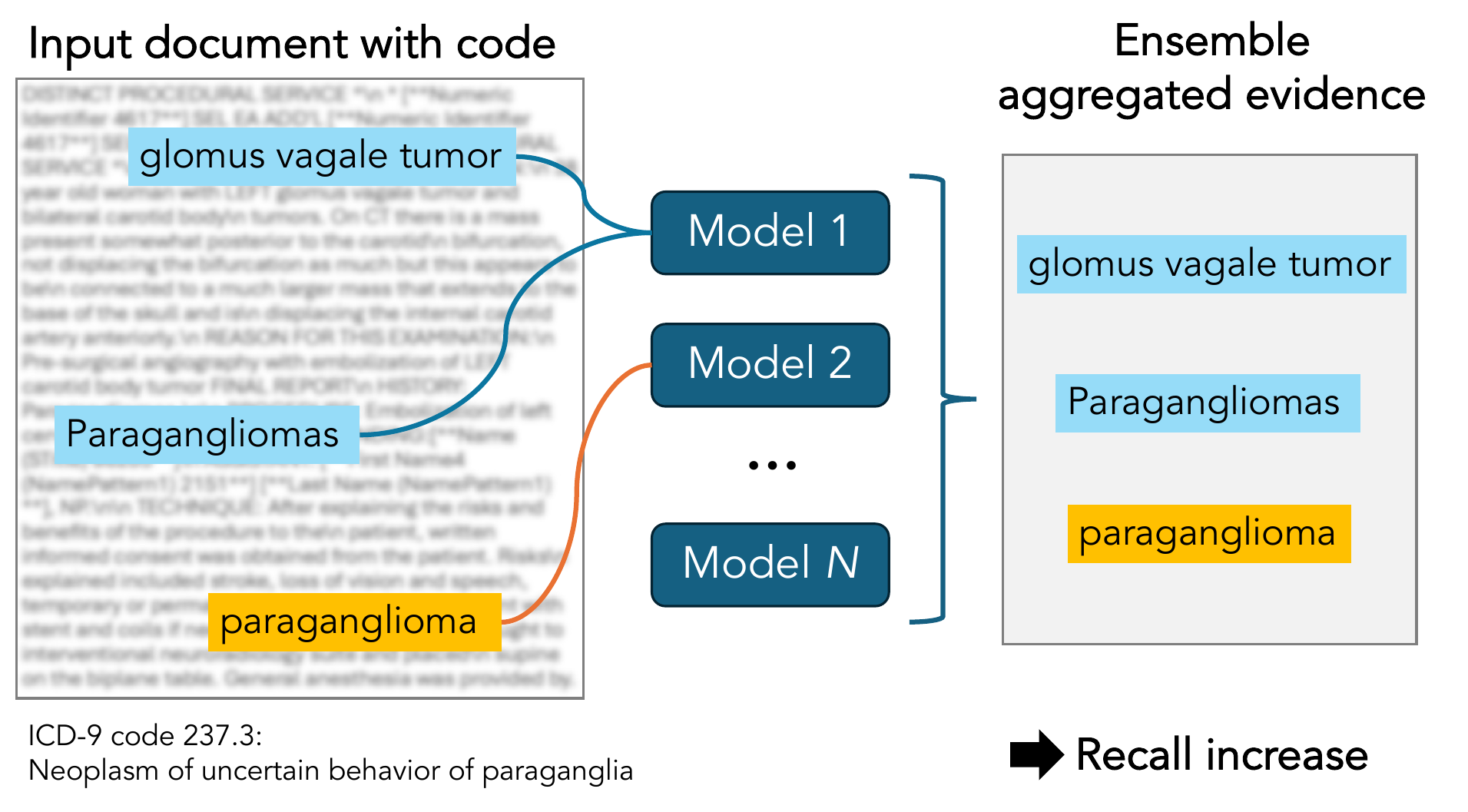}
    \caption{Illustration of complete evidence extraction on the medical coding task. An ensemble approach, aggregating the evidence of several models, leads to higher recall and, thus, more complete evidence.}
    \label{fig:fig1}
\end{figure}
While sufficient evidence extraction is model-centric and concerned with faithfully explaining model predictions, completeness is data-centric and demands the full set of cues available in the input, which is especially hard to realize in long text.
Rather than tackling this objective with a single model, we draw inspiration from work utilizing multiple models \cite{kobylinska2024exploration,cavus2025beyond,hsurashomon}: 
Different models achieve similar classification performance while relying on distinct solutions, a phenomenon referred to as the Rashomon effect~\cite{breiman2001statistical,muller2023empirical,rashomon-amazing-2024}. 
If individual models attend to different valid cues, aggregating their evidence may increase coverage of human-annotated evidence. This intuition also connects to observed benefits of diversity in clinical decision making \cite{yuan2026mixed}. 
We therefore explore whether ensembles can recover more complete evidence than single models. 

For this, we present a case study investigating complete evidence extraction for a medical coding task (\autoref{fig:fig1}). The dataset consists of long clinical notes with human-annotated evidence. Using existing domain-specific models and feature attribution scores \cite{edin-etal-2024-unsupervised}, we build a Rashomon ensemble formed by multiple equally-accurate models and aggregate evidence extracted from all models in the ensemble. 

Our contributions are as follows: 
\setlist{nolistsep}
\begin{itemize}[noitemsep]
    \item We formulate and empirically study complete evidence extraction in a medical coding task with human-annotated evidence.
    \item We demonstrate that Rashomon ensembles significantly improve evidence recall over single models with only a small token overhead.
    \item We provide empirical guidance on ensemble design: evidence-guided training improves precision but not recall; ensembles of already three models outperform any single model.   
\end{itemize}

\section{Background}
Feature attribution methods assign importance scores to input features with respect to a model prediction \cite{ribeiro2016should,sundararajan2017axiomatic}. 
\citet{li2024a} formalize completeness as the extent to which all predictive features are captured. Their study was conducted in controlled synthetic settings where predictive features are known. 
In text classification, the true predictive features are latent.
We therefore approximate them with human-annotated evidence and focus on complete evidence as introduced by \citet{cheng-etal-2023-mdace}: all text spans that support a label.\footnote{The notion of \emph{comprehensiveness} \cite{deyoung-etal-2020-eraser} is similar, but completeness as understood here does not directly assume that evidence removal leads to a reduction in model confidence.}

Operationally, we equate completeness with high recall of human-annotated complete evidence.
Prior work exploits model diversity for more trustworthy explanations or clinical decisions \citep{kobylinska2024exploration, yuan2026mixed}, but does not study completeness of textual evidence. 
Here, we study Rashomon ensembles of same-architecture language models that differ only by random seed, holding the attribution method fixed, and analyze whether aggregating evidence improves completeness in an applied setting.

\section{Methods}
We perform a case study on clinical coding investigating evidence retrieval from several models. We compare ensembles to single models measuring recall of human-annotated evidence and evidence similarity between models. The evaluation is on evidence recall, not on classification performance.   

\paragraph{Task and Dataset}
We consider the classification task of assigning medical codes to free-text clinical notes as the basis task and focus our evaluation on the newly framed complete evidence extraction task. 
MDACE \cite{cheng-etal-2023-mdace} is a medical dataset based on MIMIC-III \cite{goldberger2000physiobank,johnson2016mimic}. It contains electronic health records in the English language with clinical codes (ICD-9). For each code, evidence is provided in the form of annotated text spans, e.g., code 416.8 (other pulmonary heart disease) has `pulmonary hypertension' as one evidence span. 

MDACE documents are annotated in a sufficient and also a complete style, in which \textit{all} text that is relevant for a code is annotated. The complete subset has roughly three times more evidence. 
Discharge summaries (6,000 tokens on average) serve as data basis. 
We follow the dataset split by \citet{cheng-etal-2023-mdace} of 181 train, 60 validation, 61 test hospital admissions (586 test documents). 
In contrast to most prior work, we utilize the subset with complete evidence for testing. 
Filtering for cases with more than one evidence span yields 17 test cases for recall analysis, limiting generalizability. 
Due to the small amount of data, we additionally use the whole test set with 586 test cases (with only sufficient evidence) to analyze how stable the results are measuring evidence similarity of models and number of unique tokens.   
To our knowledge, MDACE is the only available dataset offering complete evidence annotations for long clinical text.

\paragraph{Models}
We use existing model weights from \citet{edin-etal-2024-unsupervised}. The underlying architecture is a transformer, encoder-based, trained on medical text and fine-tuned on MDACE. From the models, we selected two training regimes to investigate whether training style affects ensemble performance: 
\setlist{nolistsep}
\begin{itemize}[noitemsep]
    \item Input Gradient Regularization (IGR): no use of evidence labels, but gradient penalty to reduce irrelevant token importance.
    \item Evidence-Guided Training (EGT): uses human evidence spans as supervision to guide the model’s attention to those tokens.
\end{itemize}
Each approach contains 10 seeds from random initializations, leading to a total of 20 models.

\paragraph{Ensembles}
For each approach, IGR and EGT, Rashomon ensembles consist of the 10 respective models. 
We aggregate the evidence of all models, i.e., the union of extracted tokens. 
For the exploration of optimal ensemble size, we report scores for all possible model combinations.%\footnote{Experiments with confidence gating, keeping only evidence above a probability threshold, did not show performance improvements.}

\paragraph{Evidence Extraction}
To retrieve evidence, a feature attribution method is employed. 
We use AttInGrad scores from \citet{edin-etal-2024-unsupervised} which had the highest faithfulness and plausibility metrics on MDACE in prior work \cite{edin-etal-2024-unsupervised,beckh-etal-2025-anatomy}. The decision threshold was set based on a validation set.

\paragraph{Metrics}
Using human-annotated evidence as ground truth, we compute token-level recall
\(R = \frac{|G \cap M|}{|G|}\)
and precision
\(P = \frac{|G \cap M|}{|M|}\),
where \(G\) and \(M\) are the sets of ground-truth and predicted token IDs. Since missing evidence is costlier than checking extra tokens, we focus on recall as our completeness measure. We assess recall improvements via 10,000-fold bootstrap resampling of test cases, reporting mean recall differences between the ensemble and the best single model with percentile-based 95\% confidence intervals and Bonferroni-corrected p-values. Evidence overlap of models in ensembles is measured by pairwise Jaccard similarity.

\section{Results}
\begin{table}
  \centering
  \small
  \begin{tabularx}{\linewidth}{@{} >{\raggedright\arraybackslash}X @{\hskip 1pt} >{\raggedright\arraybackslash}X @{\hskip 1pt} *{2}{>{\raggedright\arraybackslash}X} @{}}
    \toprule
    \textbf{Train type} & \textbf{Metric} & \textbf{Single model} &  \textbf{Ensemble}  \\
    \midrule
    \multirow{2}{*}{IGR} & \textit{recall}    & 0.60 \tiny($\pm$0.25) &  0.87 \tiny($\pm$0.23)  \\
                         & \textit{precision} & 0.70 \tiny($\pm$0.21)   & 0.49 \tiny($\pm$0.18)  \\
                         
    \midrule
    \multirow{2}{*}{EGT} & \textit{recall}    & 0.63 \tiny($\pm$0.26) &  0.86 \tiny($\pm$0.24)  \\
                         & \textit{precision} & 0.74 \tiny($\pm$0.22)   & 0.57 \tiny($\pm$0.24)  \\
                         
    \bottomrule
  \end{tabularx}
  \caption{Mean recall and precision values (+ standard deviation) for single models and the Rashomon ensemble comprising 10 models, for input gradient regularization (IGR) and evidence-guided training (EGT).}
  \label{tab1}
\end{table}

\paragraph{Single model vs. ensemble} 
Table \ref{tab1} displays mean recall and precision scores for the single models and the ensemble approach under both training regimes. 
The ensembles achieve higher recall than the single models. Bootstrap resampling shows that these gains are statistically significant: 
$\Delta_{\text{IGR}} = 0.16$ (95\% CI [0.06, 0.26], $p_{\text{corr}} = 0.0036$) and $\Delta_{\text{EGT}} = 0.22$ (95\% CI [0.11, 0.32], $p_{\text{corr}} < 0.0002$).

As evidence is aggregated in the ensemble approach, precision is reduced due to the additional evidence tokens. 
When comparing the training regimes, EGT shows higher precision scores, but no recall improvement for the ensemble. Thus, using evidence cues during training may improve precision but has no notable effect on recall.  

\paragraph{Ensemble size} Figure \ref{fig:box} shows mean, minimum, and maximum recall values for each ensemble size. 
For each size, we report all possible model combinations, e.g., for size 2 we analyze 45 model pairs.
With each additional model, more evidence information is retrieved, unless it is a very low-performing combination.  
The marginal utility gains of added models decline, i.e., they are most pronounced in the beginning and reduce with increasing ensemble size. 
Adding only one model already leads to a recall increase of roughly 0.1.
IGR and EGT show similar recall patterns. 
One difference is that IGR has lower minimum values which is likely due to an outlier model with worse performance. 
From 3 models onward (4 in the case of IGR), even the lowest performing combination has higher recall than the best single model.

\begin{figure*}
    \centering
    \begin{subfigure}[t]{0.47\textwidth}
    \resizebox{!}{4.4cm}{%
    \hspace*{17pt}\input{fig/boxplot_steps_igr4.pgf}%
    }
    \caption{IGR}
    \label{fig:box_igr}
    \end{subfigure}
    \hfill
    \begin{subfigure}[t]{0.47\textwidth}
    \resizebox{!}{4.4cm}{%%
    \hspace*{17pt}\input{fig/boxplot_steps_egt4.pgf}%
    }
    \caption{EGT}
    \end{subfigure}
    \label{fig:box_egt}
    \caption{Evidence recall for different ensemble sizes for all possible model combinations. Red diamonds show mean value, whiskers indicate minimum and maximum value, i.e., worst and best performing model combination.}
    \label{fig:box}
\end{figure*}

\paragraph{Evidence similarity}
\begin{table}[t]
\centering
\small
\begin{tabularx}{\columnwidth}{@{} *{5}{>{\raggedright\arraybackslash}X} @{}}
\toprule
 & \multicolumn{2}{c}{\textbf{Evidence similarity}} & \multicolumn{2}{c}{\textbf{Unique tokens}} \\
\cmidrule(lr){2-3} \cmidrule(lr){4-5}
 & \textit{full test set} & \textit{complete test set} & \textit{full test set} & \textit{complete test set} \\
\midrule
IGR & 0.52 {\scriptsize($\pm$0.12)} & 0.52 {\scriptsize($\pm$0.15)} & 10.58 {\scriptsize($\pm$4.9)} & 11.06 {\scriptsize($\pm$3.98)} \\
\midrule
EGT & 0.55 {\scriptsize($\pm$0.06)} & 0.57 {\scriptsize($\pm$0.07)} & 10.27 {\scriptsize($\pm$5.69)} & 9.88 {\scriptsize($\pm$4.30)} \\
\bottomrule
\end{tabularx}
\caption{Evidence similarity of the 10 models in the ensemble measured by mean pairwise Jaccard similarity, number of unique evidence tokens in the ensemble on full MDACE test set and subset with complete evidence.}
\label{tab:agreement_length}
\end{table}

Table \ref{tab:agreement_length} shows evidence similarity measured by pairwise Jaccard similarity and unique token counts for the full test set and the subset with complete evidence. 
Evidence similarity is between 0.52 and 0.57 indicating a substantial evidence overlap between the models. The agreement and the number of unique tokens are similar on both test sets, confirming the same evidence patterns for more data.  

\paragraph{Qualitative insights}  
When inspecting data points, we anecdotally find that EGT models extract more clinically relevant tokens, whereas IGR models more frequently highlight function words and punctuation. 
In one case of low recall, the human-annotated evidence `glomus vagale tumor' (code 237.3) could not be reliably retrieved. 
Nearly all models identified meaningful tokens `paragangli', `vagus', and subword `omus', but none captured `tumor'.
This may be because `tumor' occurs in many codes and is a \emph{descriptive}, not a \emph{discriminative} feature. 
We also observed under-annotation in the human-annotated spans: in several cases, the ensemble discovered valid supporting spans that were not present in the annotation, suggesting imperfect human-annotated evidence (also see \citealp{khadka2025}). Thus, ensemble precision may be underestimated and higher than reported. 

\section{Discussion} 
Our main result is that aggregated evidence from Rashomon ensembles significantly increased evidence recall in a medical coding task. This validates that different models retrieve meaningful evidence that single models could not. 
The diversity in evidence patterns supports this: pairwise Jaccard similarity between models is moderate (0.52–0.57), suggesting substantial overlap in retrieved tokens but also systematic differences. Aggregating evidence across models leverages this diversity. Our findings are an empirical instance of benefiting from the Rashomon effect \cite{breiman2001statistical}.

Practically, the results imply that when exhaustive coverage is required, Rashomon ensembles are a useful strategy, provided that measures are implemented to handle the increase in false positives. 
For this task, an ensemble consisting of 10 models retrieves 10-11 tokens per document. This number is small relative to the typical length of discharge summaries (around 6,000 tokens). The overhead therefore is acceptable in environments where it is costly to miss evidence. 
Our results are in line with similar work on mixed-vendor systems \cite{yuan2026mixed}. While prior work focused on LLMs, our case study shows recall benefits on small domain-specific language models from different initializations.

\paragraph{Limitations and future work}
Here, we simply aggregate model evidence by taking the union of evidence tokens. For consensus finding, we envision more sophisticated methods which keep relevant cues while also improving precision. 
Our study focused on encoder models and a single feature attribution method, which may not be representative of the whole landscape of possible evidence extraction approaches for transformer architectures.
The analysis was limited by the amount of available data,  restricting generalizability. We addressed this by applying the ensemble approach to more test data with only sufficient evidence, showing that the same patterns of evidence similarity and number of tokens emerge. We hypothesize that this transfers to similar settings with long texts and potentially beyond: for scientific discovery, ensembles may help surface weak or rare signals. 
%For generalization, more datasets with complete annotations are needed. 
A promising direction for future work is to steer model diversity directly during training \cite{pmlr-v97-pang19a}.

\section{Conclusion}
We presented the first systematic case study of complete evidence extraction for clinical coding, focusing on recovering all relevant evidence tokens.
Aggregating evidence from multiple models significantly increased recall compared to single models with only a small token overhead. Ensembles of already three models outperform any single model, suggesting that exploiting model diversity is a suitable strategy when exhaustive coverage is required.

\paragraph*{Acknowledgments} We would like to thank Elisa Studeny for support in data processing. We are grateful to Sebastian M\"uller, Vanessa Toborek, and the NLU team for valuable discussions.

\bibliography{custom}

\end{document}